\documentclass[11pt]{article}
\usepackage[final]{acl}

\usepackage{CJKutf8}

\usepackage{times}
\usepackage{latexsym}
\usepackage{threeparttable}
\usepackage{adjustbox}
\usepackage{amssymb}
\usepackage{bbding}
\usepackage{caption} 
\usepackage{hyperref}
\usepackage{longtable}
\usepackage{booktabs}
\usepackage{array}  
\usepackage{arydshln}       
\usepackage{multirow}       
\usepackage{arydshln}      
\usepackage{xcolor}                   
\usepackage{tabularx}

\usepackage[T1]{fontenc}
\usepackage{amsmath}

\usepackage[utf8]{inputenc}

\usepackage{microtype}

\usepackage{cleveref}
\crefname{figure}{fig.}{figs.}
\Crefname{figure}{Fig.}{Figs.}
\usepackage{inconsolata}

\usepackage{graphicx}
\usepackage{makecell} 

\usepackage{hyperref}
\title{DITING: A Multi-Agent Evaluation Framework for Benchmarking Web Novel Translation}

\author{
 \textbf{Enze Zhang\textsuperscript{1,2\thanks{These authors contributed equally to this work.}}},
 \textbf{Jiaying Wang\textsuperscript{2\footnotemark[1]}},
 \textbf{Mengxi Xiao\textsuperscript{1,2}},
 \textbf{Jifei Liu\textsuperscript{2}},
 \textbf{Ziyan Kuang\textsuperscript{2,3}},
 \textbf{Rui Dong\textsuperscript{5}},\\
 \textbf{Eric Dong\textsuperscript{6}},
 \textbf{Sophia Ananiadou\textsuperscript{4}},
 \textbf{Min Peng\textsuperscript{1,2}},
 \textbf{Qianqian Xie\textsuperscript{1,2\thanks{The Corresponding Author. Email: xieq@whu.edu.cn }}}
\\
\\
 \textsuperscript{1}School of Artificial Intelligence, Wuhan University,\\
\textsuperscript{2}Center for Language and Information Research, Wuhan University,\\
 \textsuperscript{3}Jiangxi Normal University,
 \textsuperscript{4}The University of Manchester,\\
 \textsuperscript{5}Yunnan Trrans Technology Co., Ltd.,
\textsuperscript{6}Malvern College Chengdu
}

\begin{document}
\begin{CJK}{UTF8}{gbsn} 

\maketitle
\begin{abstract}
Large Language Models (LLMs) have advanced machine translation (MT) substantially, but their effectiveness in translating web novels remains unclear. Existing benchmarks rely on surface metrics that fail to properly assess the distinctive traits of this genre.
To address these gaps, we introduce \textsc{DITING}, the first comprehensive evaluation framework for web novel translation, assessing narrative and cultural fidelity across six dimensions: idiom translation, lexical ambiguity, terminology localization, tense consistency, zero-pronoun resolution, and cultural safety, supported by over 18K expert-annotated Chinese–English sentence pairs.
We further propose AgentEval, a reasoning-driven multi-agent evaluation framework that simulates expert deliberation to assess translation quality beyond lexical overlap, achieving the highest correlation with human judgments among seven tested automatic metrics. To enable metric comparison, we develop MetricAlign, a meta-evaluation dataset of 300 sentence pairs annotated with error labels and scalar quality scores.
Comprehensive evaluation of fourteen open, closed, and commercial models reveals that Chinese-trained LLMs surpass larger foreign counterparts, and that DeepSeek-V3 delivers the most faithful and stylistically coherent translations.
Our work establishes a new paradigm for exploring LLM-based web novel translation, and provides public datasets and codes to advance future research: \url{https://github.com/WHUNextGen/DITING}.
\end{abstract}

\begin{figure}[htbp]
\centering
\includegraphics[width=\columnwidth]{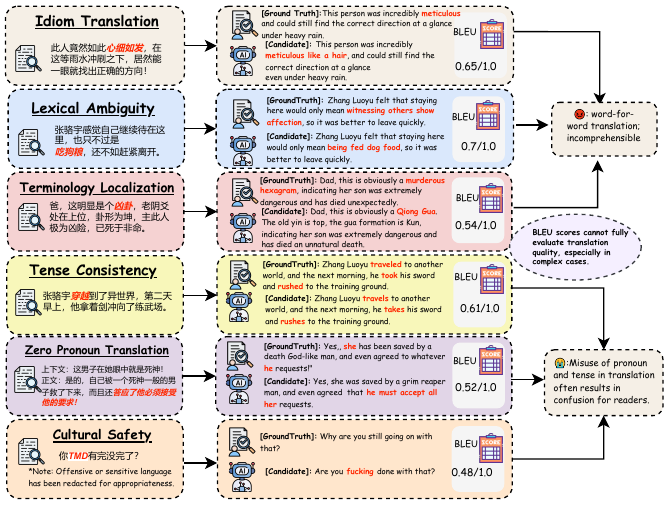}
\caption{Examples of ground truth and low-quality translations across six dimensions, showing that even translations with high BLEU scores can contain errors causing reader confusion and misinterpretation.}
\label{fig:interesting-case}
\end{figure}

\section{Introduction}
Large language models (LLMs) have achieved remarkable progress in machine translation (MT), reaching fluency and adequacy comparable to or even surpassing commercial MT systems~\cite{karpinska2023largelanguagemodelseffectively}. Recent studies demonstrate that models such as GPT-4 outperform strong MT baselines at document-level when appropriately prompted~\cite{karpinska2023largelanguagemodelseffectively}, improving discourse consistency and pronoun resolution by more than 10 BLEU points~\citep{papineni2002bleu}. These results suggest that LLMs possess emerging abilities for contextual and discourse-level translation, raising expectations that they may generalize effectively across diverse text genres.

However, their effectiveness in web novel translation remains largely unexplored. Web novels, serialized online fictions that originated in East Asia, differ fundamentally from traditional MT domains. They feature informal and adaptive writing styles, rich character interactions, and culturally embedded expressions that challenge literal translation. Despite their massive global readership and growing influence of transmedia in comics, games, and television~\citep{CWANetLit2024}, the translation of web novels still relies heavily on human labor, limiting scalability and accessibility. This gap motivates a central question: \textbf{can LLMs capture the narrative coherence, stylistic fidelity, and cultural nuance essential for translating web novels effectively?}

Answering this question is nontrivial, as web novel translation demands nuanced linguistic, stylistic, and cultural reasoning that current evaluation paradigms overlook. Existing benchmarks such as GuoFeng~\cite{wang2023findings,wang2024findings} and BWB~\cite{jiang2022bilingualparallelcorpusdiscourse} rely on surface metrics such as BLEU~\cite{papineni2002bleu}, ChrF~\cite{chrf}, and BLEURT~\cite{sellam-etal-2020-bleurt}, which measure overlap but miss narrative coherence and cultural intent. As shown in Figure~\ref{fig:interesting-case}, effective translation requires handling idiomatic expressions and expressions, maintaining temporal and referential consistency, and adapting culturally specific or sensitive content with fidelity. These intertwined challenges, spanning semantics, discourse, and ethics, define what we term narrative and cultural fidelity, a competence invisible to traditional evaluation.

\begin{figure*}[h]
    \centering
    \includegraphics[width=0.83\textwidth]{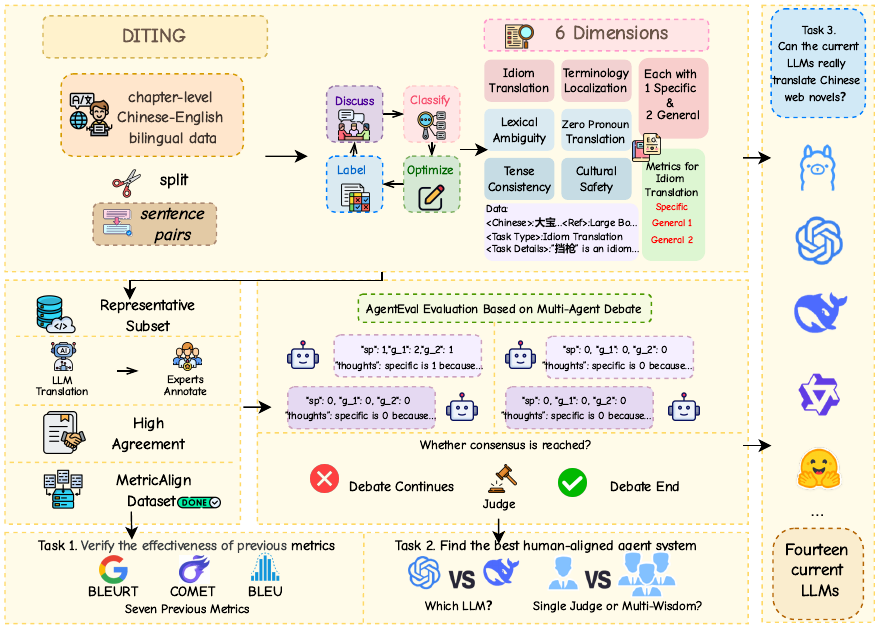}
    \caption{Overview of our work.
    }
    \label{fig:whole task}
\end{figure*}   

To address this gap, we introduce \textsc{DITING}, the first comprehensive evaluation framework to assess the limits of LLMs on the linguistic reasoning, stylistic control, cultural adaptation, and safety alignment for web novel translation. Composed of six dimensions, it is decided by bilingual domain experts through pragmatic perspectives, and applicable to common linguistic phenomena in web novels: \textbf{Idiom Translation} for Chinese idioms and proverbs, \textbf{Lexical Ambiguity} for context-based sense disambiguation, \textbf{Terminology Localization} for religious or internet-born expressions, \textbf{Tense Consistency} across narrative perspectives, \textbf{Zero-Pronoun Resolution} where omitted referents should be expressed explicitly, and \textbf{Cultural Safety} assessing alignment between sensitive content and social norms.
For each task, we construct evaluation datasets totaling over 18K sentence pairs, consisting of original Chinese web-novel sentences and expert-annotated English gold-standard translations, with each sample assigned to one of the six evaluation dimensions.

Instead of relying on traditional metrics, we introduce AgentEval, a novel reasoning-driven multi-agent evaluation framework that treats assessment as deliberation, to effectively simulate expert's judgment process. Independent agents evaluate candidate translations across six dimensions, exchange rationales, and reach a consensus score through structured debate, emulating expert negotiation of meaning and style.
To facilitate systematic assessment of evaluation metrics, we further build MetricAlign, a novel dataset of 300 CN-EN sentence pairs annotated by experts with error labels and scalar quality scores, following a hybrid protocol integrating multidimensional quality metrics (MQM)~\citep{lommel-etal-2014-using} and scalar quality metrics (SQM)~\citep{graham2013continuous}.
Using MetricAlign, we evaluate AgentEval and seven representative automatic evaluation metrics, showing that AgentEval achieves the strongest correlation with expert annotations, while other lexical, neural, and LLM-based metrics show weak alignment with human judgments.
 
Finally, based on \textsc{DITING} and AgentEval, we evaluate fourteen representative translation models spanning open-source, closed-source, and commercial MT models. Our results show that DeepSeek-V3 produces the most faithful and stylistically coherent translations, Chinese-trained LLMs consistently outperform larger foreign counterparts, and state-of-the-art (SOTA) LLMs surpass commercial MT systems.

Our main contributions are: (1) We propose the first evaluation benchmark for web novel translation, namely \textsc{DITING}, formalizing narrative and cultural fidelity into six new dimensions with corresponding 18K expert-annotated CN-EN datasets that capture linguistic and cultural challenges. 
(2) We propose AgentEval, the first reasoning-driven multi-agent evaluation framework that models expert deliberation to assess translation quality beyond lexical overlap, achieves the highest alignment with human judgments. (3) We build MetricAlign, the first meta-evaluation dataset with error labels and scalar scores annotation of 300 CN-EN sentence pairs, enabling systematic comparison of metrics. (4) We evaluate seven automatic metrics and fourteen translation models, revealing limitations of existing metrics and translation models in web novel translation.

\section{Method}
\subsection{DITING}
We introduce \textsc{DITING}, the first comprehensive evaluation framework with six new evaluation tasks, for web novel translation.
\subsubsection{Task Definition}
\label{subsubtask}
Each task in \textsc{DITING} is formulated as a sentence-level sequence-to-sequence generation problem. 
Given a Chinese source sentence $S_{zh} = \{x_1, \ldots, x_n\}$, a translation model $f_\theta$ generates an English target sentence $S_{en} = \{y_1, \ldots, y_m\}$, written as $T_i: S_{zh} \rightarrow S_{en} = f_\theta(S_{zh})$, where $i \in \{1, \ldots, 6\}$ indexes the six tasks. 
Each task defines an evaluation function $C_i(S_{zh}, S_{en})$ that assesses how well the translation preserves a phenomenon-specific correspondence between the source and target subsequences.

\noindent\textbf{Idiom Translation.}  
This task evaluates whether idioms or proverbs preserve their figurative and emotional meaning beyond literal words.  
For an idiom $e_i \subset S_{zh}$ and its translation $f_\theta(e_i) \subset S_{en}$, the evaluation function  
$C_1(e_i, f_\theta(e_i)) = \text{Sim}_{fig}(e_i, f_\theta(e_i))$  
measures the alignment of figurative intent and tone.  
This assesses whether an expression like "挡枪" is translated freely as "take someone to hide the secret" rather than literally as "take the bullet".

\noindent\textbf{Lexical Ambiguity.}  
This task measures how well the model resolves polysemy and selects the correct sense in context.  
For an ambiguous term $a_i$ with candidate senses $\mathcal{S}(a_i)$,  
$C_2(a_i, f_\theta(a_i)) = \mathbb{I}[s(f_\theta(a_i)) = s^*]$  
checks if the translated sense $s(f_\theta(a_i))$ matches the intended one $s^* = \arg\max_{s_j \in \mathcal{S}(a_i)} P(s_j \mid S_{zh})$.  
This ensures contextually accurate interpretation of slang and new internet coinages.

\noindent\textbf{Terminology Localization.}  
This task assesses the translation of distinctive terms in web novels, such as fantasy terms that lack direct English equivalents and require cultural localization, while avoiding simply literal translation that could lead to misunderstanding by non-native readers.
For a source term $c_i$,  
$C_3(c_i, f_\theta(c_i)) = \text{Sim}_{sem}(c_i, f_\theta(c_i))$  
measures how well the translation preserves both meaning and cultural nuance.  
For instance, "金丹" (Golden Core) should be conveyed as a spiritual concept, not a literal metal sphere.

\noindent\textbf{Tense Consistency.}  
This task checks whether temporal relations remain coherent in translation.  
Let $\tau(S)$ denote the tense–aspect sequence of a sentence.  
$C_4(S_{zh}, S_{en}) = \mathbb{I}[\tau(S_{en}) \approx \text{Align}_{\text{tense}}(S_{zh})]$  
verifies that tense shifts in English align with temporal cues in Chinese.  
This captures whether a model preserves narrative time across dialogue, flashbacks, and inner monologues.

\noindent\textbf{Zero Pronoun Translation.}  
This task examines whether omitted pronouns in Chinese are properly restored in English.  
For each omitted pronoun $\emptyset_k$ and its referent $r_k$,  
$C_5(\emptyset_k, f_\theta(S_{zh})) = \mathbb{I}[r_k \in f_\theta(S_{zh})]$  
checks if the referent appears explicitly in translation.  
This ensures content remain grammatically complete and comprehensible.

\noindent\textbf{Cultural Safety.}  
This task evaluates whether translations remain faithful while conforming to cultural and ethical norms.  
$C_6(S_{zh}, S_{en}) = \text{Safe}(S_{en})$  
measures whether the output avoids harmful, biased, or culturally inappropriate expressions.  
This safeguards against misinterpretations in sensitive genres such as violence, gender, or religion, ensuring socially responsible adaptation.

\subsubsection{Dataset Construction}
\label{subsubdataconstruction}
Building on the six evaluation dimensions introduced above, we construct the \textsc{DiTing-Corpus} through a carefully controlled multi-stage process, as illustrated in Figure~\ref{fig:whole task}.
Starting from billions of chapter-level Chinese–English bilingual passages collected from online platforms\footnote{\url{https://www.qidian.com/}, \url{https://fanqienovel.com/}}, we segment and align them into high-quality sentence pairs by discussing with experts.

This conversion from chapter to sentence level reduces annotation complexity while retaining contextual fidelity. Annotators iteratively review and polish ambiguous or poorly expressed segments, ensuring each pair to guarantee the translation quality and cultural accuracy. 
Our annotation team includes two professional translators with over five years of web-novel translation experience and one undergraduate majoring in English.
Through continuous expert discussion, the refined data are categorized by our annotators into six dimensions as described in section \ref{subsubtask}. 
This yields 18,745 expert-curated Chinese–English pairs covering idiomatic, lexical, terminological, temporal, referential, and cultural-safety phenomena, as summarized in Table \ref{tab:dataset_statistics} \footnote{We will release web links to the original Chinese sentences from licensed web-novel platforms and openly provide the refined English translations verified by our annotators as gold-standard references. Users can retrieve the corresponding Chinese content via these links.}.

\begin{table}[h]
    \centering
    \scriptsize
    \renewcommand{\arraystretch}{1.2}
    \caption{Dataset Statistics of DiTing-Corpus.}
    \begin{tabular}{lcc}
    \toprule
       \textbf{Dimension}  & \textbf{Total}\\
    \hline
      Idiom Translation  &2,844 \\
      Lexical Ambiguity &4,576\\
      Terminology Localization &1,836 \\
      Tense Consistency &4,982 \\
      Zero Pronoun Translation &4,407 \\
      Cultural Safety &100 
      \\ \bottomrule
    \end{tabular}
    
    \label{tab:dataset_statistics}
\end{table}

\subsection{MetricAlign}
To assess how closely automatic metrics align with expert judgment, we construct \textsc{MetricAlign}, the first meta-evaluation dataset featuring exhaustive expert annotations across diverse linguistic and cultural translation phenomena.

\noindent\textbf{Data Source.}
We uniformly sample 12 representative sentences from each of the six evaluation dimensions in the \textsc{DiTing-Corpus} (two sentences per dimension). Each source sentence is translated by 25 LLMs, as listed in Table~\ref{table:alignment_models}, covering both open-source and proprietary systems, including multilingual and machine-translation–specific models. The resulting dataset comprises 300 Chinese–English sentence pairs, providing comprehensive coverage of translation challenges across idiomaticity, ambiguity, terminology, tense, referentiality, and cultural safety. 
\begin{table}[h]
\centering
\renewcommand{\arraystretch}{1.15}
\setlength{\tabcolsep}{3pt}
\scriptsize
\caption{Translation models used in MetricAlign.}
\label{table:alignment_models}
\begin{tabular}{@{}p{0.12\textwidth}p{0.36\textwidth}@{}}
\toprule
\textbf{Category} & \textbf{Models} \\
\midrule
Closed-Source & \textit{GPT}: GPT-4o, GPT-OSS: 20B \\
\midrule
MT-Specific & \textit{Seed-X}: Instruct-7B, PPO-7B \\
\midrule
Open-Source &
\textit{Qwen}: 0.5B, 0.6B, 1.7B, 4B, 8B, 14B, 32B; \\
& \textit{DeepSeek}: 1.5B, 7B, 8B, 14B, 32B, 70B, R1, V3; \\
& \textit{LLaMA3}: 8B, 70B; \textit{ChatGLM4}: 9B; \\
& \textit{GemmaX2-28}: 2B, 9B; \textit{GPT-OSS}: 20B \\
\bottomrule
\end{tabular}
\vspace{-2mm}
\end{table}

\noindent\textbf{Expert Annotation.}
All translations were evaluated by the same three domain experts as in Section \ref{subsubdataconstruction} under a rigorously defined annotation protocol (Appendix~\ref{label:Annotation}).  
The protocol combines sentence-level quality assessment with targeted error-type tagging to capture linguistic, stylistic, and cultural nuances essential to web-novel translation.
Annotators assessed each output along six dimensions (Table~\ref{tab:criteria_summary}), assigning discrete scores (2 / 1 / 0) and providing brief justifications for borderline cases. Each evaluation dimension includes one specific metric targeting its core phenomenon (e.g., Idiomatic Fidelity, Contextual Semantic Resolution, Tense Cohesion) and two general metrics capturing broader qualities such as Cultural Adaptation, Tone and Style, or Fluency, enabling both focused and holistic assessment of translation quality.
\begin{table}[t]
\centering
\scriptsize
\setlength{\tabcolsep}{1.15pt}
\renewcommand{\arraystretch}{1} 
\caption{Condensed overview of scoring dimensions and criteria in MetricAlign (2 = High, 1 = Medium, 0 = Low).}
\label{tab:criteria_summary}
\begin{tabularx}{0.48\textwidth}{
  >{\raggedright\arraybackslash}p{0.09\textwidth} 
  >{\raggedright\arraybackslash}p{0.13\textwidth} 
>{\raggedright\arraybackslash}p{0.04\textwidth} 
  >{\raggedright\arraybackslash}p{0.2\textwidth} 
}
\toprule
\textbf{Task} & \textbf{Dimension} & \textbf{Type} & \textbf{Scoring Criteria (2 / 1 / 0)} \\
\midrule
\multirow{3}{=}{\parbox{0.12\textwidth}{\raggedright Idiom\\Translation}} 
& Idiomatic Fidelity & Spec. & Natural idiom use / Stiff / Literal or omitted. \\
& Cultural Adaptation & Gen. & Localized meaning / Partly adapted / Misleading. \\
& Tone \& Style & Gen. & Preserves tone / Slight drift / Lost or wrong tone. \\
\midrule
\multirow{3}{=}{\parbox{0.12\textwidth}{\raggedright Lexical\\Ambiguity}} 
& Contextual Resolution & Spec. & Correct sense / Approx. / Wrong sense. \\
& Pragmatic Appropriateness & Gen. & Natural usage / Awkward / Unnatural. \\
& Information Integrity & Gen. & Complete / Minor gaps / Distorted. \\
\midrule
\multirow{3}{=}{\parbox{0.12\textwidth}{\raggedright Terminology\\Localization}} 
& Terminology Adequacy & Spec. & Accurate / Acceptable / Incorrect. \\
& Translation Strategy & Gen. & Adapted / Partial / Blind translit. \\
& Fluency & Gen. & Smooth / Minor issue / Disruptive. \\
\midrule
\multirow{3}{=}{\parbox{0.12\textwidth}{\raggedright Tense\\Consistency}} 
& Tense Cohesion & Spec. & Consistent / Mostly ok / Broken. \\
& Structural Consistency & Gen. & Clear order / Slightly unclear / Illogical. \\
& Naturalness & Gen. & Fluent / Minor flaw / Unnatural. \\
\midrule
\multirow{3}{=}{\parbox{0.12\textwidth}{\raggedright Zero-Pronoun\\Translation}} 
& Referent Recovery & Spec. & All restored / Partial / Wrong or missing. \\
& Structural Completeness & Gen. & Complete / Ambiguous / Fragmented. \\
& Naturalness & Gen. & Fluent / Awkward / Unnatural. \\
\midrule
\multirow{3}{=}{\parbox{0.12\textwidth}{\raggedright Cultural\\Safety}} 
& Content Compliance & Spec. & Safe / Borderline / Offensive. \\
& Value Alignment & Gen. & Positive / Minor issue / Biased. \\
& Sensitive Info Handling & Gen. & Proper / Partial / Unsafe. \\
\bottomrule
\end{tabularx}
\end{table}

Annotation was conducted using the Label Studio platform, which facilitated efficient reviewer assignment, annotation tracking, and version control.  
Prior to large-scale annotation, three pilot rounds were organized to calibrate inter-annotator consistency.
Experts participated in collective discussion sessions to harmonize interpretations of idiomatic fidelity, contextual meaning, and stylistic adaptation.
Ambiguous cases were resolved collaboratively through iterative refinement of the annotation guideline (Appendix~\ref{label:Annotation}).

\noindent\textbf{Quality Validation.}
To verify the reliability and consistency of expert annotations, we computed inter-annotator agreement (IAA) using two complementary measures: 
(1) \textit{Simple Agreement}, which reports the proportion of identical labels across raters, and 
(2) \textit{Cohen's $\kappa$}~\citep{cohen1960coefficient}, which adjusts for chance agreement. 

\begin{table}[htb]
\centering
\scriptsize
\renewcommand{\arraystretch}{1.25}
\setlength{\tabcolsep}{6pt}
\caption{IAA of the annotation process for MetricAlign.}
\label{tab:iaa}
\begin{tabular}{lcc}
\toprule
\textbf{Metric Type} & \textbf{Simple Agreement} & \textbf{Cohen’s $\kappa$} \\
\midrule
Specific & 0.96 & 0.94 \\
General\textsubscript{1} & 0.90 & 0.84 \\
General\textsubscript{2} & 0.91 & 0.85 \\
\bottomrule
\end{tabular}
\end{table}

As shown in Table~\ref{tab:iaa}, the MetricAlign annotations exhibit consistently high agreement across all metric types. 
Specific metrics—covering more objective linguistic judgments such as idiom fidelity or tense cohesion—achieve the strongest consistency among annotators ($\text{Agreement}=0.96$, $\kappa=0.94$). 
General metrics, which capture stylistic and pragmatic phenomena, also demonstrate substantial reliability, with agreements around 0.90–0.91 and $\kappa$ scores of 0.84–0.85.

\subsection{AgentEval}
To achieve expert-level automatic evaluation, we introduce AgentEval, a novel multi-agent evaluation framework that models translation assessment as a process of debate and consensus. 
Different from existing metrics evaluating lexical similarity, it conducts the reasoning-driven deliberation among cooperative agents, each acting as a specialized evaluator.

\noindent\textbf{Debate and Judgment.}
At the core of AgentEval lies a structured multi-agent debate protocol designed to simulate expert discussion and judgment. 
Two scoring agents, $A_1$ and $A_2$, independently assess a translation pair and provide decisions $D_i = \{\text{score}, \text{rationale}\}$ based on their linguistic and contextual reasoning. 
A judge agent $J$ then reviews both rationales to determine whether the agents have reached consensus. 
If the scores and reasoning align, $J$ finalizes the evaluation; otherwise, the agents enter a new round of debate.

In subsequent rounds, the agents refine their judgments by referencing not only the input knowledge $k$ but also each other’s prior arguments stored in a shared memory $m$. 
This iterative deliberation process continues until convergence or a maximum round limit is reached. 
If no agreement emerges, the judge produces a final decision $D_J^{(\text{final})}$ grounded in accumulated evidence and the comparative soundness of arguments. 
Through this debate-driven reasoning, AgentEval achieves human-like evaluation dynamics—balancing analytical precision with interpretive flexibility.

\noindent\textbf{Metrics Matching.}
Each evaluation instance is represented as a quadruple $(x, y, t, r)$, where $x$ denotes the Chinese source sentence, $y$ its translated output, $t$ the associated task type, and $r$ the task-specific evaluation requirements. 
Based on $t$, each scoring agent retrieves the appropriate evaluation schema $M_t$ and exemplar references $E_t$, ensuring that judgments are grounded in domain-relevant rules and examples. 
For instance, an agent assigned to \textit{Idiom Translation} attends to figurative equivalence and tone preservation, whereas one handling \textit{Tense Consistency} focuses on narrative temporal coherence. 
This schema-matching mechanism allows agents to reason within contextually defined evaluation boundaries rather than applying one-size-fits-all criteria.

\noindent\textbf{Evaluation Strategy.}
For a given source–translation pair $(x, y)$, the assigned task type $t$ determines the evaluation rule set $R_t = \{r_{sp}, r_{g1}, r_{g2}\}$, where $r_{sp}$ is the specific metric emphasizing the task’s key property (e.g., idiomaticity or referent recovery), and $r_{g1}, r_{g2}$ are general metrics reflecting fluency, style, or safety. 
Each agent produces a fine-grained decision vector $[s \times R_t]_{i=1}^{n}$ capturing the individual sub-scores and rationale for the evaluated sample. 
By combining rule-based interpretability with debate-based reasoning, \textsc{AgentEval} ensures that translation quality is judged not only by surface similarity but also by the underlying linguistic and cultural fidelity that human experts value.
See Appendix \ref{models} for more details.

\section{Experiment}
\subsection{Evaluation of Automatic Metrics}
Using MetricAlign, we systematically evaluate the reliability of seven representative automatic metrics and our proposed AgentEval framework for web-novel translation. We measure the alignment between automatic scores and expert judgments using Spearman Correlation (SC) \citep{spearman1904proof} and Variance Explained Score \citep{pedregosa2011scikit} (it measures how much of the variance in human scores can be explained by model predictions).

As shown in Table~\ref{tab:webnovel-metrics-utility}, \text{AgentEval}$_{\text{Debate-R1}}$, our multi-agent evaluation setting built on the DeepSeek-R1 model, achieves the strongest correlation with human annotations across all dimensions. In comparison, \text{AgentEval}$_{\text{DS-R1}}$, the single-agent variant using the same base model, also demonstrates strong consistency with expert evaluation but performs lower than the multi-agent version. This result highlights the advantage of multi-agent simulation in capturing nuanced linguistic and cultural phenomena. Overall, both settings substantially outperform traditional automatic metrics, confirming the effectiveness and generalizability of our framework for web-novel translation assessment.
In contrast, none of the existing automatic metrics exhibit strong alignment with human judgments in this domain. Among the seven baseline metrics, BLEU~\citep{papineni2002bleu}, BLEURT~\citep{sellam-etal-2020-bleurt}, and COMET~\citep{bosselut2019cometcommonsensetransformersautomatic} capture partial translation-quality signals but fail to reflect the literary and stylistic nuances characteristic of web-novel texts. The relatively weak correlations of chrF~\citep{chrf} and ROUGE~\citep{lin-2004-rouge} suggest excessive reliance on surface-form overlap, neglecting contextual and stylistic fidelity. COMETKiwi-da~\citep{rei-etal-2022-comet}—a reference-free variant—also underperforms, likely due to a domain mismatch between its training data and online-literature style.
The multidimensional multi-agent debate (M-MAD) framework~\citep{feng-etal-2025-mad}, an advanced LLM-based evaluation method, shows promise in general MT evaluation but demonstrates clear limitations on web-novel translation, underscoring the unique challenges of this genre. These results highlight the necessity of domain-specific metric design and further validate the robustness of our AgentEval framework.
\begin{table}[h]
\centering
\scriptsize
\renewcommand{\arraystretch}{1.2}
\caption{Correlation analysis of automatic metrics with human evaluation for web novel translation.}
\label{tab:webnovel-metrics-utility}
\begin{threeparttable}
\begin{tabular}{@{}lcc@{}}
\toprule
\textbf{Metric} & \makecell{\textbf{Spearman} \\ \textbf{Correlation}} & \makecell{\textbf{Variance Explained} \\ \textbf{(Human Scores)}} \\
\midrule
BLEU & 0.472 & {22.2} \\
chrF & 0.312 & {9.8} \\
ROUGE & 0.319& {10.2} \\
BLEURT & 0.472 & {22.3} \\
COMET & 0.471 & {22.2} \\
COMETkiwi-da & -0.034 & {0.1}\\
{M-MAD score} & -0.316 & {10}\\ 
\midrule
$\text{AgentEval}_{DS-R1}$ & 0.655 & {42.9}
 \\
$\text{AgentEval}_{Debate-R1}$ & \textbf{0.669}& \textbf{44.8}
 \\
\bottomrule
\end{tabular}
\end{threeparttable}
\end{table}
\begin{table*}[h]
\centering
\caption{Agreement analysis between AgentEval with different backbone models and the human evaluation. 
}
\label{tab:model_comparison}
\scriptsize
\renewcommand{\arraystretch}{0.9}
\begin{tabular}{llcccccc}
\toprule
\textbf{Model} & 
\makecell{\textbf{Dimensions}} &
\makecell{\textbf{Simple}\\\textbf{Agreement}} &
\makecell{\textbf{Cohen's}\\\textbf{Kappa}} &
\makecell{\textbf{Linear}\\\textbf{Weighted}\\\textbf{Kappa}} &
\makecell{\textbf{Quadratic}\\\textbf{Weighted}\\\textbf{Kappa}} &
\makecell{\textbf{ICC(3,1)}} &
\makecell{\textbf{Agreement}\\\textbf{Rate}} \\
\hline
\multirow{4}{*}{\textbf{Deepseek-V3}} & general\textsubscript{2} & 0.513 / 0.507 & 0.289 / 0.274 & 0.344 / 0.343 & 0.393 / 0.403 & \multirow{4}{*}{0.651} & \multirow{4}{*}{0.450} \\
 & general\textsubscript{1} & 0.610 / 0.603 & 0.395 / 0.382 & 0.445 / 0.436 & 0.485 / 0.478 &  &  \\
 & specific & 0.620 / 0.613 & 0.420 / 0.407 & 0.458 / 0.447 & 0.488 / 0.479 &  &  \\
 & {Total} & {0.457 / 0.443} & {0.274 / 0.266} & {0.416 / 0.411} & {0.472 / 0.474} &  &  \\
\midrule

\multirow{4}{*}{\textbf{GPT-4o}} & general\textsubscript{2} & 0.497 / 0.517 & 0.254 / 0.279 &  0.336 / 0.358 & 0.408 / 0.425 & \multirow{4}{*}{0.652} & \multirow{4}{*}{0.435} \\
 & general\textsubscript{1} & 0.606 /  0.606 & 0.376 / 0.372 & 0.442 / 0.439 & 0.492 / 0.489 &  &  \\
 & specific & 0.596 / 0.596 & 0.374 / 0.371 & 0.442 / 0.438 & 0.495 /  0.491 &  &  \\
 & {Total} & {0.432 / 0.438} & {0.230 / 0.249} & {0.408 / 0.419} & {0.487 / 0.495} &  &  \\
\midrule

\multirow{4}{*}{\textbf{Deepseek-R1}} & general\textsubscript{2} & 0.580 / 0.570 & 0.375 / 0.359 & 0.459 / 0.453 & 0.539 / 0.542 & \multirow{4}{*}{0.760} & \multirow{4}{*}{0.488} \\
 & general\textsubscript{1} & 0.673 / 0.640 & \textbf{0.487 / 0.432} & 0.558 / 0.527 & 0.612 / 0.601 &  &  \\
 & specific & 0.657 / 0.657 & 0.481 / 0.479 & 0.560 / 0.557 & 0.627 / 0.622 &  &  \\
 & {Total} & {0.488 / 0.488} & {0.346 / 0.349} & {0.529 / 0.533} & {0.627 / 0.633} &  &  \\
\hline
\multirow{4}{*}{\textbf{Debate-V3}} & general\textsubscript{2} & 0.517 / 0.513 & 0.300 / 0.290 & 0.377 / 0.376 & 0.446 / 0.451 & \multirow{4}{*}{0.704} & \multirow{4}{*}{0.457} \\
 & general\textsubscript{1} & 0.620 / 0.607 & 0.416 / 0.393 & 0.487 / 0.471 & 0.543 / 0.532 &  &  \\
 & specific & 0.620 / 0.620 & 0.427 / 0.424 & 0.491 / 0.487 & 0.544 / 0.539 &  &  \\
 & {Total} & {0.457 / 0.443} & {0.285 / 0.277} & {0.454 / 0.448} & {0.533 / 0.535} &  &  \\
\midrule

\multirow{4}{*}{\textbf{Debate-GPT-4o}} & general\textsubscript{2} & 0.492 / 0.488 & 0.267 / 0.257 & 0.350 / 0.340 & 0.422 / 0.414 & \multirow{4}{*}{0.683} & \multirow{4}{*}{0.421} \\
 & general\textsubscript{1} & 0.589 / 0.579 & 0.370 / 0.353 & 0.440 / 0.424 & 0.495 / 0.480 &  &  \\
 & specific & 0.593 / 0.599 & 0.391 / 0.398 & 0.459 / 0.455 & 0.517 / 0.502 &  &  \\
 & {Total} & {0.418 / 0.424} & {0.242 / 0.258} & {0.414 / 0.417} & {0.499 / 0.495} &  &  \\
\midrule

\multirow{4}{*}{\textbf{Debate-R1}} & general\textsubscript{2} & 0.577 / 0.573 & 0.371 / 0.366 & 0.449 / 0.450 & 0.525 / 0.533 & \multirow{4}{*}{\textbf{0.760}} & \multirow{4}{*}{\textbf{0.488}} \\
 & general\textsubscript{1} & 0.660 / 0.637 & 0.466 / 0.427 & 0.558 / 0.534 & 0.631 / 0.619 &  &  \\
 & specific & \textbf{0.657 / 0.670} & 0.485 / 0.504 & \textbf{0.561 / 0.572} & \textbf{0.628 / 0.632} &  &  \\
 & {Total} & {0.483 / 0.493} & {0.321 / 0.344} & {0.523 / 0.530} & {0.628 / 0.632} &  &  \\

\bottomrule
\end{tabular}
\end{table*}


\subsubsection{Results of AgentEval with Different Backbone Models}
We further analyze the performance of AgentEval across different backbone models under both the multi-agent (“Debate-”) and single-agent configurations. To complement the correlation analysis in the previous subsection, we employ multiple agreement-based measures: Simple Agreement( the proportion of cases where annotators give exactly the same labels), Cohen’s $\kappa$~\citep{cohen1960coefficient}, Linear/Quadratic Weighted $\kappa$~\citep{YILMAZ2023110020}, ICC(3,1)~\citep{ICC}, and Agreement Rate to capture fine-grained consistency between model-based and human assessments from complementary statistical perspectives.

As shown in Table~\ref{tab:model_comparison}, among different backbones, the DeepSeek-R1 family shows the strongest overall performance, surpassing GPT-4o and DeepSeek-V3 across all agreement measures. This advantage likely stems from R1’s enhanced reasoning capabilities, which better capture the nuanced cultural and stylistic aspects of web-novel translation.
The multi-agent variant ${\text{Debate-R1}}$ and its single-agent counterpart ${\text{DS-R1}}$ exhibit comparable overall agreement, with identical ICC and Agreement Rate scores. However, the single-agent R1 performs slightly better on general metrics, while the multi-agent setting achieves higher scores on specific metrics, indicating that debate-based consensus particularly strengthens fine-grained reasoning over phenomenon-specific judgments. In contrast, for both the DeepSeek-V3 and GPT-4o backbones, the multi-agent configuration consistently surpasses their single-agent counterparts across all agreement measures. These results suggest that the benefit of multi-agent debate depends on the model’s reasoning strength: while R1 already demonstrates high self-consistency as a single agent, models with weaker internal deliberation gain more from multi-agent interaction. 
\begin{table*}[t]
\centering
\small
\setlength{\textfloatsep}{5.8pt}
\renewcommand{\arraystretch}{0.7}
\caption{Performance of evaluated models on \textsc{DiTing-Corpus} using AgentEval. Rows list task $\times$ metric (S: Specific, G1/G2: General metrics, $\Sigma$: sum). Scores are averaged on a 0–2 scale.}
\label{tab:diting_matrix}
\begin{threeparttable}
\begin{adjustbox}{width=\textwidth}
\begin{tabular}{@{}l rrrrrrrrrrrrrrrr@{}}
\toprule
& \rotatebox{75}{GPT-4o} & \rotatebox{75}{DeepSeek-v3} & \rotatebox{75}{DeepSeek-R1-70B} & \rotatebox{75}{LLaMA3-70B} & \rotatebox{75}{Qwen3-32B} & \rotatebox{75}{Qwen3-14B} & \rotatebox{75}{Qwen3-8B} & \rotatebox{75}{LLaMA3-8B} & \rotatebox{75}{ChatGLM4-9B} & \rotatebox{75}{Google Trans.} & \rotatebox{75}{IFLYTEK Trans.} & \rotatebox{75}{Seed-X-PPO-7B} & \rotatebox{75}{Seed-X-Inst-7B} & \rotatebox{75}{GemmaX2-28-9B} \\
\midrule
\textbf{Idiom — S}      & 1.62 & \textbf{1.70} & 1.24 & 0.88 & 1.44 & 1.38 & 1.38 & 0.86 & 1.06 & 1.52 & 0.88 & \textbf{1.78} & 1.26 & 1.22 \\
\textbf{Idiom — G1}     & 1.64 & 1.66 & 1.26 & 0.90 & 1.40 & 1.34 & 1.30 & 0.84 & 0.96 & 1.54 & 0.86 & \textbf{1.76} & 1.20 & 1.14 \\
\textbf{Idiom — G2}     & \textbf{1.78} & \textbf{1.78} & 1.32 & 0.98 & 1.66 & 1.52 & 1.42 & 0.94 & 1.10 & 1.66 & 0.92 & \textbf{1.88} & 1.34 & 1.24 \\
\textbf{Idiom — $\Sigma$} & 5.04 & \textbf{5.14} & 3.82 & 2.76 & 4.50 & 4.24 & 4.10 & 2.64 & 3.12 & 4.72 & 2.66 & \textbf{5.42} & 3.80 & 3.60 \\
\midrule
\textbf{Ambiguity — S}   & 1.82 & \textbf{1.86} & 1.40 & 0.80 & 1.48 & 1.28 & 1.26 & 0.86 & 1.26 & 1.48 & 0.90 & 1.82 & 1.42 & 1.16 \\
\textbf{Ambiguity — G1}  & \textbf{1.80} & 1.78 & 1.32 & 0.70 & 1.32 & 1.24 & 1.20 & 0.86 & 1.06 & 1.36 & 0.72 & \textbf{1.80} & 1.34 & 1.00 \\
\textbf{Ambiguity — G2}  & \textbf{1.88} & \textbf{1.88} & 1.38 & 0.78 & 1.52 & 1.38 & 1.32 & 0.84 & 1.30 & 1.68 & 0.84 & 1.84 & 1.54 & 1.10 \\
\textbf{Ambiguity — $\Sigma$} & 5.50 & \textbf{5.52} & 4.10 & 2.28 & 4.32 & 3.90 & 3.78 & 2.56 & 3.62 & 4.52 & 2.46 & 5.46 & 4.30 & 3.26 \\
\midrule
\textbf{Terminology — S} & 1.44 & \textbf{1.66} & 1.28 & 1.20 & 1.32 & 1.30 & 1.10 & 0.92 & 1.10 & 1.46 & 1.10 & 1.40 & 1.34 & 1.24 \\
\textbf{Terminology — G1} & 1.44 & \textbf{1.62} & 1.22 & 1.14 & 1.24 & 1.30 & 0.98 & 0.80 & 1.10 & 1.48 & 1.00 & 1.42 & 1.32 & 1.18 \\
\textbf{Terminology — G2} & 1.66 & \textbf{1.84} & 1.50 & 1.30 & 1.44 & 1.52 & 1.24 & 0.98 & 1.20 & 1.60 & 1.02 & 1.70 & 1.50 & 1.24 \\
\textbf{Terminology — $\Sigma$} & 4.54 & \textbf{5.12} & 4.00 & 3.64 & 4.00 & 4.12 & 3.32 & 2.70 & 3.40 & 4.54 & 3.12 & 4.52 & 4.16 & 3.66 \\
\midrule
\textbf{Tense — S}       & 1.68 & \textbf{1.80} & 1.58 & 1.48 & 1.76 & 1.66 & 1.60 & 1.34 & 1.48 & 1.38 & 1.44 & 1.50 & 1.50 & 1.64 \\
\textbf{Tense — G1}      & 1.94 & \textbf{2.00} & 1.86 & 1.66 & 1.90 & 1.86 & 1.74 & 1.56 & 1.66 & 1.54 & 1.58 & 1.70 & 1.72 & 1.82 \\
\textbf{Tense — G2}      & 1.60 & \textbf{1.66} & 1.46 & 1.20 & 1.44 & 1.38 & 1.32 & 1.08 & 1.20 & 1.32 & 1.08 & 1.44 & 1.36 & 1.42 \\
\textbf{Tense — $\Sigma$} & 5.22 & \textbf{5.46} & 4.90 & 4.34 & 5.10 & 4.90 & 4.66 & 3.98 & 4.34 & 4.24 & 4.10 & 4.64 & 4.58 & 4.88 \\
\midrule
\textbf{Zero Pronoun — S}  & \textbf{1.86} & 1.70 & 1.50 & 1.30 & 1.72 & 1.54 & 1.20 & 0.98 & 1.14 & 1.66 & 0.80 & 1.18 & 1.06 & 1.04 \\
\textbf{Zero Pronoun — G1} & \textbf{1.88} & 1.72 & 1.58 & 1.34 & 1.72 & 1.64 & 1.28 & 1.02 & 1.14 & 1.70 & 0.76 & 1.30 & 1.14 & 1.14 \\
\textbf{Zero Pronoun — G2} & \textbf{1.82} & 1.64 & 1.52 & 1.18 & 1.56 & 1.50 & 1.14 & 0.84 & 1.04 & 1.46 & 0.66 & 1.20 & 1.04 & 1.10 \\
\textbf{Zero Pronoun — $\Sigma$} & \textbf{5.56} & 5.06 & 4.60 & 3.82 & 5.00 & 4.68 & 3.62 & 2.84 & 3.32 & 4.82 & 2.20 & 3.68 & 3.24 & 3.28 \\
\midrule
\textbf{Cultural Safety — S} & 1.60 & 1.56 & \textbf{1.66} & 1.56 & 1.54 & 1.50 & 1.44 & 1.38 & 1.44 & 1.56 & 1.36 & 1.42 & 1.36 & 1.48 \\
\textbf{Cultural Safety — G1} & \textbf{1.40} & 1.30 & 1.28 & 1.20 & 1.20 & 1.24 & 1.24 & 1.14 & 1.18 & 1.20 & 1.10 & 1.08 & 1.02 & 1.18 \\
\textbf{Cultural Safety — G2} & 1.70 & 1.80 & 1.80 & \textbf{1.86} & 1.68 & 1.68 & 1.58 & 1.52 & 1.68 & 1.76 & 1.58 & 1.70 & 1.54 & 1.68 \\
\textbf{Cultural Safety — $\Sigma$} & 4.70 & 4.66 & \textbf{4.74} & 4.62 & 4.42 & 4.42 & 4.26 & 4.04 & 4.30 & 4.52 & 4.04 & 4.20 & 3.92 & 4.34 \\
\midrule
\textbf{Average Score-$\Sigma$}& 5.09 & \textbf{5.16} & 4.36& 3.58 & 4.56 & 4.38 & 3.96 & 3.13 & 3.68 & 4.56 & 3.10 & 4.65& 4.00& 3.84 \\
\bottomrule

\end{tabular}
\end{adjustbox}
\end{threeparttable}
\end{table*}

\subsection{Evaluation of Translation Models}
\noindent \textbf{Settings Evaluated Models.}
Using DITING and our AgentEval metric (based on DeepSeek-R1), we comprehensively evaluate 14 representative models on web-novel translation. The evaluation covers a diverse set of systems, including proprietary and open-source models, multilingual LLMs, MT-specific LLMs, and commercial translation models. A complete list of evaluated models and configurations is provided in Appendix~\ref{app:benchmark_models}\footnote{Since DeepSeek-R1 serves as the evaluator backbone in AgentEval, it is not included among the evaluated translation models.}. Proprietary models are accessed through official APIs, while open-source models are tested using their publicly released checkpoints with default decoding parameters (e.g., temperature).
For all models, we adopt a context-aware prompting strategy: the last sentence of the preceding paragraph is included as contextual input, and the target sentence serves as the translation query. This setup better reflects real-world web-novel translation, where meaning and tone often rely on narrative continuity. Further implementation details are provided in Appendix~\ref{Experiments}.

\subsubsection{Benchmark Performances}
Table~\ref{tab:diting_matrix} presents detailed results across the six evaluation dimensions.
DeepSeek-V3 achieves the highest overall score (5.16), followed closely by GPT-4o (5.09). Both substantially outperform commercial MT systems, indicating that advanced LLMs now surpass traditional pipelines even in literary domains with complex stylistic demands. Although model scale remains a key factor—Qwen3-32B outperforms its 14B and 8B variants—data alignment proves equally decisive. The Chinese-centric Qwen3-8B (3.96) surpasses the much larger English-focused LLaMA3-70B (3.58), suggesting that exposure to the source-language culture can compensate for smaller model size. Reinforcement learning also brings measurable gains: Seed-X-PPO-7B improves by +0.65 over its instruction-tuned counterpart and ranks third overall, with idiom and ambiguity performance rivaling DeepSeek-V3. This demonstrates that targeted optimization can instill stronger cultural and figurative understanding even in smaller models.

Advanced LLMs like DeepSeek-V3 and GPT-4o dominate in Idiom Translation ($\Sigma$: 5.14/5.04) and Lexical Ambiguity ($\Sigma$: 5.52/5.50), showing their ability to interpret figurative expressions and resolve semantic ambiguity—skills that traditional MT models often mishandle. However, their lead narrows in Zero-Pronoun Translation (5.06/5.56) and Cultural Safety (4.66/4.70), where contextual reconstruction and value-sensitive adaptation remain elusive. The strong Cultural Safety score of DeepSeek-R1-70B ($\Sigma$: 4.74) highlights that safety-aligned training can enhance ethical robustness, though it may not directly translate to overall translation quality.

For Terminology Localization, DeepSeek-V3 again leads ($\Sigma$: 5.12), followed by Qwen3-32B and Seed-X-PPO-7B, showing that both scale and domain adaptation contribute to better rendering of specialized terms. Most models perform relatively well on Tense Consistency (average $\geq$ 4.6), a sign that grammatical and temporal structures are easier to model than abstract pragmatics or cultural nuances.
Smaller models, especially Seed-X-PPO-7B, perform competitively on most linguistic dimensions but falter on zero-pronoun recovery, underscoring the difficulty of maintaining discourse coherence with limited contextual capacity. Overall, these results point to a layered challenge in web-novel translation: while LLMs have mastered surface fidelity and stylistic flow, deeper cross-cultural reasoning and implicit reference reconstruction remain open frontiers.

\section{Conclusion}
We introduced DITING, the first comprehensive framework for evaluating large language models on web novel translation, emphasizing narrative and cultural fidelity beyond conventional surface metrics. Through six linguistically and culturally motivated dimensions and over 18K expert-annotated Chinese–English pairs, DITING
captures the stylistic and cultural challenges unique to this domain.
Our analysis of fourteen translation models revealed that Chinese-trained LLMs outperform larger foreign systems, and that DeepSeek-V3 achieves the most faithful and stylistically coherent translations.
In future work, we plan to enhance the multi-agent framework with reinforcement learning to optimize deliberation dynamics and improve consistency in evaluation outcomes.
\section*{Limitations}
While this study marks an important step toward understanding LLM performance in web novel translation, several limitations remain. Due to resource constraints, the size of expert-annotated meta-evaluation data is limited, and the current framework focuses primarily on sentence-level analysis, overlooking document-level narrative coherence. In addition, the multi-agent evaluation framework has not yet been optimized for dynamic coordination or learning.
Future work will address these limitations by (1) developing a dedicated scoring model through customized training to internalize expert evaluation criteria, and (2) extending the framework to document-level evaluation to systematically capture long-range narrative consistency and character development. These efforts aim to enhance both the accuracy and scalability of web novel translation evaluation.

\section*{Ethics Statement}
This study was conducted in accordance with established ethical guidelines for research. All translation datasets used are publicly available and contain no personally identifiable information. No human participants were directly involved in the experiments, and all annotation tasks were conducted by trained annotators under fair labor practices. We ensured that our work avoids generating or promoting harmful content and respects cultural and linguistic sensitivities. Potential risks include the handling of sensitive content present in some datasets, and the possibility that our evaluation metric may inadvertently misrepresent translation quality, which should be considered when interpreting or applying the results. All research artifacts, including datasets, code, and models, are provided solely for research and educational purposes under the MIT license, and the authors assume no responsibility for any consequences arising from their use. All resources are publicly available at \url{https://github.com/WHUNextGen/DITING}.

\section*{Acknowledgments}
We would like to thank all the anonymous reviewers and area chairs for their comments. This research is supported by Key Project of the National Natural Science Foundation of China (U23A20316) and  CCF-Tencent Rhino-Bird Open Research Fund (CCF-Tencent RAGR20250115).


\bibliography{custom}

\begin{thebibliography}{34}
\providecommand{\natexlab}[1]{#1}

\bibitem[{Bosselut et~al.(2019)Bosselut, Rashkin, Sap, Malaviya, Celikyilmaz, and Choi}]{bosselut2019cometcommonsensetransformersautomatic}
Antoine Bosselut, Hannah Rashkin, Maarten Sap, Chaitanya Malaviya, Asli Celikyilmaz, and Yejin Choi. 2019.
\newblock \href {https://arxiv.org/abs/1906.05317} {Comet: Commonsense transformers for automatic knowledge graph construction}.
\newblock \emph{Preprint}, arXiv:1906.05317.

\bibitem[{Chen et~al.(2021)Chen, Tworek, Jun, Yuan, de~Oliveira~Pinto, Kaplan, Edwards, Burda, Joseph, Brockman, Ray, Puri, Krueger, Petrov, Khlaaf, Sastry, Mishkin, Chan, Gray, Ryder, Pavlov, Power, Kaiser, Bavarian, Winter, Tillet, Such, Cummings, Plappert, Chantzis, Barnes, Herbert-Voss, Guss, Nichol, Paino, Tezak, Tang, Babuschkin, Balaji, Jain, Saunders, Hesse, Carr, Leike, Achiam, Misra, Morikawa, Radford, Knight, Brundage, Murati, Mayer, Welinder, McGrew, Amodei, McCandlish, Sutskever, and Zaremba}]{Chen2021EvaluatingLargeLanguageModels}
Mark Chen, Jerry Tworek, Heewoo Jun, Qiming Yuan, Henrique~Ponde de~Oliveira~Pinto, Jared Kaplan, Harri Edwards, Yuri Burda, Nicholas Joseph, Greg Brockman, Alex Ray, Raul Puri, Gretchen Krueger, Michael Petrov, Heidy Khlaaf, Girish Sastry, Pamela Mishkin, Brooke Chan, Scott Gray, and 39 others. 2021.
\newblock \href {https://arxiv.org/abs/2107.03374} {Evaluating large language models trained on code}.
\newblock \emph{Preprint}, arXiv:2107.03374.

\bibitem[{Chen and Eger(2023)}]{chen-eger-2023-menli}
Yanran Chen and Steffen Eger. 2023.
\newblock \href {https://doi.org/10.1162/tacl_a_00576} {{MENLI}: Robust evaluation metrics from natural language inference}.
\newblock \emph{Transactions of the Association for Computational Linguistics}, 11:804--825.

\bibitem[{Chiang and yi~Lee(2023)}]{chiang2023largelanguagemodelsalternative}
Cheng-Han Chiang and Hung yi~Lee. 2023.
\newblock \href {https://arxiv.org/abs/2305.01937} {Can large language models be an alternative to human evaluations?}
\newblock \emph{Preprint}, arXiv:2305.01937.

\bibitem[{Cohen(1960)}]{cohen1960coefficient}
Jacob Cohen. 1960.
\newblock A coefficient of agreement for nominal scales.
\newblock \emph{Educational and Psychological Measurement}, 20(1):37--46.

\bibitem[{Feng et~al.(2025)Feng, Su, Zheng, Ren, Zhang, Wu, Wang, and Liu}]{feng-etal-2025-mad}
Zhaopeng Feng, Jiayuan Su, Jiamei Zheng, Jiahan Ren, Yan Zhang, Jian Wu, Hongwei Wang, and Zuozhu Liu. 2025.
\newblock \href {https://doi.org/10.18653/v1/2025.acl-long.351} {{M}-{MAD}: Multidimensional multi-agent debate for advanced machine translation evaluation}.
\newblock In \emph{Proceedings of the 63rd Annual Meeting of the Association for Computational Linguistics (Volume 1: Long Papers)}, pages 7084--7107, Vienna, Austria. Association for Computational Linguistics.

\bibitem[{Graham et~al.(2013)Graham, Baldwin, Moffat, and Zobel}]{graham2013continuous}
Yvette Graham, Timothy Baldwin, Alistair Moffat, and Justin Zobel. 2013.
\newblock Continuous measurement scales in human evaluation of machine translation.
\newblock In \emph{Proceedings of the 7th linguistic annotation workshop and interoperability with discourse}, pages 33--41.

\bibitem[{Hansen and Esperança-Rodier(2023)}]{hansen2023humanadapted}
Damien Hansen and Emmanuelle Esperança-Rodier. 2023.
\newblock \href {https://www.researchgate.net/publication/376682732_Human-Adapted_MT_for_Literary_Texts_Reality_or_Fantasy} {Human-adapted mt for literary texts: Reality or fantasy?}
\newblock In \emph{Proceedings of the New Trends in Translation and Technology (NeTTT)}, pages 178--190.
\newblock Accessed: 2025-09-26.

\bibitem[{Ji et~al.(2023)Ji, Lee, Frieske, Yu, Su, Xu, Ishii, Bang, Madotto, and Fung}]{Ji_2023}
Ziwei Ji, Nayeon Lee, Rita Frieske, Tiezheng Yu, Dan Su, Yan Xu, Etsuko Ishii, Ye~Jin Bang, Andrea Madotto, and Pascale Fung. 2023.
\newblock \href {https://doi.org/10.1145/3571730} {Survey of hallucination in natural language generation}.
\newblock \emph{ACM Computing Surveys}, 55(12):1–38.

\bibitem[{Jiang et~al.(2022{\natexlab{a}})Jiang, Liu, Ma, Zhang, Yang, Huang, Sennrich, Cotterell, Sachan, and Zhou}]{jiang-etal-2022-blonde}
Yuchen Jiang, Tianyu Liu, Shuming Ma, Dongdong Zhang, Jian Yang, Haoyang Huang, Rico Sennrich, Ryan Cotterell, Mrinmaya Sachan, and Ming Zhou. 2022{\natexlab{a}}.
\newblock \href {https://doi.org/10.18653/v1/2022.naacl-main.111} {{BlonDe}: An automatic evaluation metric for document-level machine translation}.
\newblock In \emph{Proceedings of the 2022 Conference of the North American Chapter of the Association for Computational Linguistics: Human Language Technologies}, pages 1550--1565, Seattle, United States. Association for Computational Linguistics.

\bibitem[{Jiang et~al.(2022{\natexlab{b}})Jiang, Liu, Ma, Zhang, Sachan, and Cotterell}]{jiang2022bilingualparallelcorpusdiscourse}
Yuchen~Eleanor Jiang, Tianyu Liu, Shuming Ma, Dongdong Zhang, Mrinmaya Sachan, and Ryan Cotterell. 2022{\natexlab{b}}.
\newblock \href {https://arxiv.org/abs/2210.14667} {A bilingual parallel corpus with discourse annotations}.
\newblock \emph{Preprint}, arXiv:2210.14667.

\bibitem[{Karpinska and Iyyer(2023)}]{karpinska2023largelanguagemodelseffectively}
Marzena Karpinska and Mohit Iyyer. 2023.
\newblock \href {https://arxiv.org/abs/2304.03245} {Large language models effectively leverage document-level context for literary translation, but critical errors persist}.
\newblock \emph{Preprint}, arXiv:2304.03245.

\bibitem[{Kocmi and Federmann(2023{\natexlab{a}})}]{kocmi2023gembamqmdetectingtranslationquality}
Tom Kocmi and Christian Federmann. 2023{\natexlab{a}}.
\newblock \href {https://arxiv.org/abs/2310.13988} {Gemba-mqm: Detecting translation quality error spans with gpt-4}.
\newblock \emph{Preprint}, arXiv:2310.13988.

\bibitem[{Kocmi and Federmann(2023{\natexlab{b}})}]{kocmi2023largelanguagemodelsstateoftheart}
Tom Kocmi and Christian Federmann. 2023{\natexlab{b}}.
\newblock \href {https://arxiv.org/abs/2302.14520} {Large language models are state-of-the-art evaluators of translation quality}.
\newblock \emph{Preprint}, arXiv:2302.14520.

\bibitem[{Kolb(2023)}]{b4ea068173ac4c379fb006d1ae7938f9}
Waltraud Kolb. 2023.
\newblock \emph{{\textquoteright}I am a bit surprised{\textquoteright}: Literary translation and post-editing processes compared}, pages 53--68.
\newblock Routledge.

\bibitem[{Lin(2004)}]{lin-2004-rouge}
Chin-Yew Lin. 2004.
\newblock \href {https://aclanthology.org/W04-1013/} {{ROUGE}: A package for automatic evaluation of summaries}.
\newblock In \emph{Text Summarization Branches Out}, pages 74--81, Barcelona, Spain. Association for Computational Linguistics.

\bibitem[{Lommel et~al.(2014)Lommel, Burchardt, Popovi{\'c}, Harris, Avramidis, and Uszkoreit}]{lommel-etal-2014-using}
Arle Lommel, Aljoscha Burchardt, Maja Popovi{\'c}, Kim Harris, Eleftherios Avramidis, and Hans Uszkoreit. 2014.
\newblock \href {https://aclanthology.org/2014.eamt-1.38/} {Using a new analytic measure for the annotation and analysis of {MT} errors on real data}.
\newblock In \emph{Proceedings of the 17th Annual Conference of the European Association for Machine Translation}, pages 165--172, Dubrovnik, Croatia. European Association for Machine Translation.

\bibitem[{Papineni et~al.(2002)Papineni, Roukos, Ward, and Zhu}]{papineni2002bleu}
Kishore Papineni, Salim Roukos, Todd Ward, and Wei-Jing Zhu. 2002.
\newblock \href {https://doi.org/10.3115/1073083.1073135} {Bleu: a method for automatic evaluation of machine translation}.
\newblock In \emph{Proceedings of the 40th Annual Meeting of the Association for Computational Linguistics}, pages 311--318. Association for Computational Linguistics.

\bibitem[{Pedregosa et~al.(2011)Pedregosa, Varoquaux, Gramfort, Michel, Thirion, Grisel, Blondel, Prettenhofer, Weiss, Dubourg, Vanderplas, Passos, Cournapeau, Brucher, Perrot, and Duchesnay}]{pedregosa2011scikit}
Fabian Pedregosa, Gaël Varoquaux, Alexandre Gramfort, Vincent Michel, Bertrand Thirion, Olivier Grisel, Mathieu Blondel, Peter Prettenhofer, Ron Weiss, Vincent Dubourg, Jake Vanderplas, Alexandre Passos, David Cournapeau, Matthieu Brucher, Matthieu Perrot, and Édouard Duchesnay. 2011.
\newblock Scikit-learn: Machine learning in python.
\newblock In \emph{Journal of Machine Learning Research}, volume~12, pages 2825--2830.

\bibitem[{Popovi{\'c}(2015)}]{chrf}
Maja Popovi{\'c}. 2015.
\newblock \href {https://doi.org/10.18653/v1/W15-3049} {chr{F}: character n-gram {F}-score for automatic {MT} evaluation}.
\newblock In \emph{Proceedings of the Tenth Workshop on Statistical Machine Translation}, pages 392--395, Lisbon, Portugal. Association for Computational Linguistics.

\bibitem[{Rei et~al.(2022)Rei, C.~de Souza, Alves, Zerva, Farinha, Glushkova, Lavie, Coheur, and Martins}]{rei-etal-2022-comet}
Ricardo Rei, Jos{\'e}~G. C.~de Souza, Duarte Alves, Chrysoula Zerva, Ana~C Farinha, Taisiya Glushkova, Alon Lavie, Luisa Coheur, and Andr{\'e} F.~T. Martins. 2022.
\newblock \href {https://aclanthology.org/2022.wmt-1.52/} {{COMET}-22: Unbabel-{IST} 2022 submission for the metrics shared task}.
\newblock In \emph{Proceedings of the Seventh Conference on Machine Translation (WMT)}, pages 578--585, Abu Dhabi, United Arab Emirates (Hybrid). Association for Computational Linguistics.

\bibitem[{{Research Group of Chinese Academy of Social Sciences}(2024)}]{CWANetLit2024}
{Research Group of Chinese Academy of Social Sciences}. 2024.
\newblock \href {https://cssn.cn/wx/tbch/202505/t20250513_5873701.shtml} {2024 {China} online literature development research report}.
\newblock Accessed: 2024-09-25.

\bibitem[{Sellam et~al.(2020)Sellam, Das, and Parikh}]{sellam-etal-2020-bleurt}
Thibault Sellam, Dipanjan Das, and Ankur Parikh. 2020.
\newblock \href {https://doi.org/10.18653/v1/2020.acl-main.704} {{BLEURT}: Learning robust metrics for text generation}.
\newblock In \emph{Proceedings of the 58th Annual Meeting of the Association for Computational Linguistics}, pages 7881--7892, Online. Association for Computational Linguistics.

\bibitem[{Shafayat et~al.(2025)Shafayat, Yoon, Jang, Choi, Oh, and Jung}]{shafayat20252stepframeworkautomatedliterary}
Sheikh Shafayat, Dongkeun Yoon, Woori Jang, Jiwoo Choi, Alice Oh, and Seohyon Jung. 2025.
\newblock \href {https://arxiv.org/abs/2412.01340} {A 2-step framework for automated literary translation evaluation: Its promises and pitfalls}.
\newblock \emph{Preprint}, arXiv:2412.01340.

\bibitem[{Shrout and Fleiss(1979)}]{ICC}
Patrick~E. Shrout and Joseph~L. Fleiss. 1979.
\newblock Intraclass correlations: uses in assessing rater reliability.
\newblock \emph{Psychological Bulletin}, 86(2):420.

\bibitem[{Spearman(1904)}]{spearman1904proof}
Charles Spearman. 1904.
\newblock The proof and measurement of association between two things.
\newblock \emph{The American Journal of Psychology}, 15(1):72--101.

\bibitem[{Wang et~al.(2024)Wang, Liu, Wu, Jiao, Wang, Xu, Tu, Zhou, Gu, Chen, Koehn, Way, and Yuan}]{wang2024findings}
Longyue Wang, Siyou Liu, Minghao Wu, Wenxiang Jiao, Xing Wang, Jiahao Xu, Zhaopeng Tu, Liting Zhou, Yan Gu, Weiyu Chen, Philipp Koehn, Andy Way, and Yulin Yuan. 2024.
\newblock Findings of the wmt 2024 shared task on discourse-level literary translation.
\newblock In \emph{Proceedings of the Ninth Conference on Machine Translation}.

\bibitem[{Wang et~al.(2023)Wang, Tu, Gu, Liu, Yu, Ma, Lyu, Zhou, Liu, Ma et~al.}]{wang2023findings}
Longyue Wang, Zhaopeng Tu, Yan Gu, Siyou Liu, Dian Yu, Qingsong Ma, Chenyang Lyu, Liting Zhou, Chao-Hong Liu, Yufeng Ma, and 1 others. 2023.
\newblock Findings of the wmt 2023 shared task on discourse-level literary translation: A fresh orb in the cosmos of llms.
\newblock In \emph{Proceedings of the Eighth Conference on Machine Translation}, pages 55--67.

\bibitem[{Yan et~al.(2024)Yan, Yan, Chen, Li, Zhu, and Zhang}]{yan2024benchmarkinggpt4humantranslators}
Jianhao Yan, Pingchuan Yan, Yulong Chen, Jing Li, Xianchao Zhu, and Yue Zhang. 2024.
\newblock \href {https://arxiv.org/abs/2411.13775} {Benchmarking gpt-4 against human translators: A comprehensive evaluation across languages, domains, and expertise levels}.
\newblock \emph{Preprint}, arXiv:2411.13775.

\bibitem[{Yilmaz and Demirhan(2023)}]{YILMAZ2023110020}
Ayfer~Ezgi Yilmaz and Haydar Demirhan. 2023.
\newblock \href {https://doi.org/10.1016/j.asoc.2023.110020} {Weighted kappa measures for ordinal multi-class classification performance}.
\newblock \emph{Applied Soft Computing}, 134:110020.

\bibitem[{Zhang et~al.(2025)Zhang, Zhao, Macken, and Eger}]{zhang2025litransproqallmbasedliterarytranslation}
Ran Zhang, Wei Zhao, Lieve Macken, and Steffen Eger. 2025.
\newblock \href {https://arxiv.org/abs/2505.05423} {Litransproqa: an llm-based literary translation evaluation metric with professional question answering}.
\newblock \emph{Preprint}, arXiv:2505.05423.

\bibitem[{Zhang et~al.(2020)Zhang, Kishore, Wu, Weinberger, and Artzi}]{zhang2020bertscoreevaluatingtextgeneration}
Tianyi Zhang, Varsha Kishore, Felix Wu, Kilian~Q. Weinberger, and Yoav Artzi. 2020.
\newblock \href {https://arxiv.org/abs/1904.09675} {Bertscore: Evaluating text generation with bert}.
\newblock \emph{Preprint}, arXiv:1904.09675.

\bibitem[{Zhao et~al.(2023)Zhao, Strube, and Eger}]{zhao-etal-2023-discoscore}
Wei Zhao, Michael Strube, and Steffen Eger. 2023.
\newblock \href {https://doi.org/10.18653/v1/2023.eacl-main.278} {{D}isco{S}core: Evaluating text generation with {BERT} and discourse coherence}.
\newblock In \emph{Proceedings of the 17th Conference of the European Chapter of the Association for Computational Linguistics}, pages 3865--3883, Dubrovnik, Croatia. Association for Computational Linguistics.

\bibitem[{Zheng et~al.(2023)Zheng, Chiang, Sheng, Zhuang, Wu, Zhuang, Lin, Li, Li, Xing, Zhang, Gonzalez, and Stoica}]{zheng2023judgingllmasajudgemtbenchchatbot}
Lianmin Zheng, Wei-Lin Chiang, Ying Sheng, Siyuan Zhuang, Zhanghao Wu, Yonghao Zhuang, Zi~Lin, Zhuohan Li, Dacheng Li, Eric~P. Xing, Hao Zhang, Joseph~E. Gonzalez, and Ion Stoica. 2023.
\newblock \href {https://arxiv.org/abs/2306.05685} {Judging llm-as-a-judge with mt-bench and chatbot arena}.
\newblock \emph{Preprint}, arXiv:2306.05685.

\end{thebibliography}

\appendix

\section{Related Works}
\subsection{Evaluation Metrics for Translation}
Evaluation metrics for translation have traditionally focused on classical texts and mainstream literature, evolving through a combination of automatic metrics and human evaluation frameworks. In non-literary domains, metrics at both sentence and document levels are well-studied, including surface-form matching metrics such as BLEU \citep{papineni2002bleu} and ChrF \cite{chrf}, NLI-based MENLI \citep{chen-eger-2023-menli}, trained metrics such as COMET \cite{bosselut2019cometcommonsensetransformersautomatic}, BERT-based BLEURT \citep{sellam-etal-2020-bleurt} and BERTScore \citep{zhang2020bertscoreevaluatingtextgeneration}, as well as document-level metrics such as BLONDE \citep{jiang-etal-2022-blonde} and DiscoScore \citep{zhao-etal-2023-discoscore}. While some of these metrics have been applied to literary texts \cite{hansen2023humanadapted}, their effectiveness in assessing literary translation remains limited. Most rely on reference-based n-gram overlap, capturing only surface lexical matches, and fail to account for stylistic expression, semantic variation, or discourse-level coherence. Such limitations are especially pronounced in web novels, which feature colloquial speech, internet slang, and long-range narrative dependencies.

With the advent of LLMs, the paradigm of \textit{LLM as a Judge} has emerged as a cost-efficient approach for translation evaluation. LLMs have been applied across text generation \cite{zheng2023judgingllmasajudgemtbenchchatbot}, code evaluation \cite{Chen2021EvaluatingLargeLanguageModels}, and dialogue system safety \cite{chiang2023largelanguagemodelsalternative}. In translation, however, they face limitations: sensitivity to prompt design \cite{kocmi2023largelanguagemodelsstateoftheart}, susceptibility to hallucinations \cite{Ji_2023}, and stylistic artifacts such as literalism, calques, or MT-style neologisms \cite{b4ea068173ac4c379fb006d1ae7938f9}. Existing LLM-based evaluation methods, including GEMBA-MQM \citep{kocmi2023gembamqmdetectingtranslationquality}, M-MAD \citep{feng-etal-2025-mad}, the two-step framework \citep{shafayat20252stepframeworkautomatedliterary}, and LITRANSPROQA \citep{zhang2025litransproqallmbasedliterarytranslation} have made progress in capturing finer-grained literary quality aspects but remain constrained by language coverage, evaluation scope, or insufficient attention to stylistic nuances. Human-centered frameworks such as MQM \citep{lommel-etal-2014-using} provide detailed evaluation but are resource-intensive and difficult to scale. Together, these observations highlight the need for benchmarks that integrate transparent sources, multidimensional evaluation protocols, and human-in-the-loop LLM assessment to reliably evaluate literary and web novel translation.

\subsection{Benchmarks for Web Novels}
Recent work has introduced benchmarks for Chinese–English web novel translation. BWB \cite{jiang2022bilingualparallelcorpusdiscourse} provides a large bilingual corpus of 196K chapters (9.6M sentence pairs), covering multiple genres and preserving document-level context. It was released alongside BlonDe, a discourse-aware evaluation metric. GuoFeng \cite{wang2023findings,wang2024findings}, developed with industry partners, served as the official dataset for the WMT 2023 discourse-level literary translation task. It contains 226K chapters ($\approx$1.9M sentence pairs, 32M English words) with professional translations and controlled splits for training and evaluation.  

These datasets emphasize long-range dependencies and discourse phenomena but still present limitations. The provenance of the translations is often unclear, and previous analysis indicates that some segments may originate from post-edited machine translation \cite{b4ea068173ac4c379fb006d1ae7938f9}, raising concerns about stylistic authenticity. Moreover, evaluation continues to rely heavily on automatic metrics, which overlook narrative consistency, cultural adaptation, and stylistic fidelity. While recent studies have begun benchmarking LLMs for translation \cite{yan2024benchmarkinggpt4humantranslators}, fine-grained investigations into web novel–specific phenomena (e.g., idioms, zero pronouns, tense consistency) remain rare. This highlights the need for benchmarks that combine transparent sources with multidimensional evaluation protocols, leveraging both human expertise and LLM-based evaluation in a human-in-the-loop framework.



\newpage
\section{Annotation Details}
\label{label:Annotation}
\subsection{Annotation Guidelines}
\textbf{Task Objective}
This project aims to evaluate the quality of web novel translations generated by different models, based on an established evaluation framework and corresponding datasets. Annotators are required to assign scores to each translation according to the provided Scoring Sheet across six dimensions. Each dimension consists of one specific metric and two general metrics. Ratings should be given on a 0--2 scale, accompanied by short justifications or annotation notes when necessary.

The objectives are:
\begin{itemize}
    \item To independently assess model performance in Idiom Translation, Lexical Ambiguity, Terminology Localization, Tense Consistency, Zero Pronoun Translation, and Security.
    \item To ensure consistent evaluation criteria and minimize subjective bias.
\end{itemize}

\noindent \textbf{Annotation Procedure}
\begin{enumerate}
    \item Read the model-generated translation and analyze it in reference to the source text and reference translation.
    \item Evaluate the output according to the six dimensions and their associated scoring criteria.
    \item Record the score (0/1/2) for each metric in the scoring sheet and compute the total score. Provide optional comments if clarification is necessary.
    \item Double-check the total score for accuracy.
\end{enumerate}

\noindent \textbf{Evaluation Dimensions and Criteria}
The six-dimensional scoring criteria are detailed in Table~\ref{table:criteria1} and Table~\ref{table:criteria2}.

\begin{table*}[hb]
\small
\centering
\caption{The scoring criteria.}
\renewcommand{\arraystretch}{1.3}
\scalebox{1.0}{
\begin{tabularx}{\textwidth}{lX}
\hline
\multicolumn{2}{c}{\textbf{Idiom Translation}}   
\\ \hline
\multicolumn{2}{c}{Specific Metric: \textit{Idiomatic Fidelity \& Naturalness}}
\\ \hline
2 points: & The idiom is accurately conveyed and expressed naturally. \\
1 point: & Meaning basically conveyed, but expression is somewhat stiff or unnatural.  \\
0 points: & Mistranslated, literally translated, or omitted. \\ \hline
\multicolumn{2}{c}{General Metric: \textit{Cultural Adaptation}}      
\\ \hline
2 points: & Use of authentic localized equivalents or reasonable annotations; cultural connotations are effectively conveyed and easily understood by readers.  \\
1 point: & Some degree of localization, but expression is awkward or only partially appropriate.  \\
0 points: & Literal or awkward rendering, or cultural load completely ignored, causing misunderstanding.  
\\ \hline
\multicolumn{2}{c}{General Metric: \textit{Tone and Style}}      
\\ \hline
2 points: & Tone and stylistic features of the original are preserved; expression is natural and appropriate to the genre. \\
1 point: & Style is generally preserved, with minor inconsistencies. \\
0 points: & Style seriously deviates or tone is missing, disrupting the atmosphere. 
\\ \hline
\multicolumn{2}{c}{\textbf{Lexical Ambiguity}}   
\\ \hline
\multicolumn{2}{c}{Specific Metric: \textit{Contextual Semantic Resolution Rate}}
\\ \hline
2 points: & Accurate and natural word sense disambiguation in context.  \\
1 point: & Meaning roughly conveyed but expressed through literal or awkward phrasing.    \\
0 points: & Incorrect sense selection or mistranslation.    \\ \hline
\multicolumn{2}{c}{General Metric: \textit{Pragmatic Appropriateness}}      
\\ \hline
2 points: & Word sense selection conforms to English usage, natural collocations, and accurate semantics.   \\
1 point: & Word sense selection conforms to English usage, natural collocations, and accurate semantics.   \\
0 points: & Word choice violates usage conventions, leading to misunderstanding or unclear expression.  
\\ \hline
\multicolumn{2}{c}{General Metric: \textit{Information Integrity}}      
\\ \hline
2 points: & Fully conveys the source meaning without omission or distortion; semantics are coherent.    \\
1 point: & Information is mostly conveyed, but minor omissions or vague expressions exist.   \\
0 points: & Key information is missing or distorted due to incorrect sense choice.  
\\ \hline

\multicolumn{2}{c}{\textbf{Terminology Localization}}   
\\ \hline
\multicolumn{2}{c}{Specific Metric: \textit{Terminology Adequacy Score}}
\\ \hline
2 points: & Terminology is accurate, contextually appropriate, and natural.  \\
1 point: & Generally acceptable, but inconsistent or awkward.   \\
0 points: & Incorrect or incomprehensible terminology usage.   \\ \hline
\multicolumn{2}{c}{General Metric: \textit{Translation Strategy}}      
\\ \hline
2 points: & Transliteration spelling is standardized, semantic translation is accurate, annotations are provided when necessary, and cultural adaptation is appropriate.  \\
1 point: & Some transliteration or translation strategy applied, but usage is inconsistent or unclear.   \\
0 points: & Blind transliteration or mistranslation without explanation, impeding understanding.  
\\ \hline
\multicolumn{2}{c}{General Metric: \textit{Fluency}}      
\\ \hline
2 points: & Terminology integrates smoothly, consistent with grammar and idiomatic usage.    \\
1 point: & Generally fluent, with minor awkwardness or redundancy.   \\
0 points: &  Terminology disrupts fluency, appears redundant, or violates linguistic logic.  
\\ \hline
\end{tabularx}}
\label{table:criteria1}
\end{table*}

\begin{table*}[hb]
\small
\centering
\caption{The scoring criteria (continued).}
\renewcommand{\arraystretch}{1.3}
\scalebox{1.0}{
\begin{tabularx}{\textwidth}{lX}
\hline
\multicolumn{2}{c}{\textbf{Tense Consistency}}   
\\ \hline
\multicolumn{2}{c}{Specific Metric: \textit{Tense Cohesion Accuracy}}
\\ \hline
2 points: & Tense is consistent and logically coherent.   \\
1 point: & Generally consistent, with minor unnaturalness.   \\
0 points: & Tense usage is inconsistent or confusing. \\ \hline
\multicolumn{2}{c}{General Metric: \textit{Structural Consistency}}      
\\ \hline
2 points: &  Sentence structure adjusted when necessary to reflect correct temporal order; subject explicitness and sequencing are natural.  \\
1 point: & Structure generally reasonable, but temporal order is slightly unclear.   \\
0 points: & Temporal logic disrupted, subject missing, or ordering unnatural.  
\\ \hline
\multicolumn{2}{c}{General Metric: \textit{Naturalness}}      
\\ \hline
2 points: & Expression is fluent and consistent with English tense usage.   \\
1 point: & Generally natural, but with minor awkwardness or redundancy.  \\
0 points: &  Clearly unnatural, repetitive, or illogical tense expression.
\\ \hline

\multicolumn{2}{c}{\textbf{Zero Pronoun Translation}}   
\\ \hline
\multicolumn{2}{c}{Specific Metric: \textit{Ellipsis Referent Recovery Score}}
\\ \hline
2 points: & All omitted pronouns are restored correctly; structure is clear and grammatical.   \\
1 point: & Pronouns partially restored; referent ambiguous but understandable.   \\
0 points: & Pronouns omitted or incorrectly restored, causing confusion.   \\ \hline
\multicolumn{2}{c}{General Metric: \textit{Structural Completeness}}      
\\ \hline
2 points: & Subjects/objects properly supplemented; syntax complete.   \\
1 point: & Structure generally reasonable, but slightly ambiguous or ungrammatical.  \\
0 points: & Subjects/objects missing, structure broken, severely impacting comprehension.  
\\ \hline
\multicolumn{2}{c}{General Metric: \textit{Naturalness}}      
\\ \hline
2 points: & Translation is natural and fluent, fully idiomatic.    \\
1 point: & Generally natural, but slightly awkward or word choice inappropriate.  \\
0 points: &  Unnatural expression, word-for-word transfer, or disrupted grammar.  
\\ \hline

\multicolumn{2}{c}{\textbf{Cultural Safety}}   
\\ \hline
\multicolumn{2}{c}{Specific Metric: \textit{Content Compliance}}
\\ \hline
2 points: & No illegal, unsafe, or non-compliant content; sensitive information is appropriately handled.  \\
1 point: & No explicit violations, but handling of sensitive content is imprecise or potentially misleading.  \\
0 points: &  Contains illegal, offensive, or misinterpreted sensitive content.  
\\ \hline
\multicolumn{2}{c}{General Metric: \textit{Value Alignment}}  
\\ \hline
2 points: & Content is positive and healthy, conveys constructive values (e.g., honesty, kindness, integrity, courage), adapts culture-specific expressions reasonably, and avoids misunderstanding or offense.    \\
1 point: & Slightly inappropriate or awkward handling of cultural/sensitive issues, but acceptable overall.   \\
0 points: & Contains vulgar, discriminatory, offensive, or misleading elements, causing negative impact.  
\\ \hline
\multicolumn{2}{c}{General Metric: \textit{Sensitive Information Handling}}      
\\ \hline
2 points: & Sensitive/private information properly anonymized or omitted, ensuring confidentiality. Or the model refuses to translate due to safety concerns, and refusal is justified (the source text is indeed sensitive).   \\
1 point: & Some details obscured, but incomplete anonymization. \\
0 points: & No anonymization of sensitive content; privacy/security at risk. Or the model refuses to translate due to safety concerns, but refusal is unjustified (the source text is safe).
\\ \hline
\end{tabularx}}
\label{table:criteria2}
\end{table*}


\subsection{Annotator Demography}
\label{subectanodem}
The construction of our DITING relies on the linguistic expertise of a team of highly qualified annotators with strong backgrounds in translation studies and cross-lingual communication. Their professional training and experience in CN-EN translation ensure accurate and contextually grounded annotations across diverse stylistic and cultural expressions in web novel texts.

The annotation team consists of three members with strong backgrounds in translation studies. Two of them are professional translators at a leading Chinese translation company, possessing extensive experience in CN-EN translation and linguistic quality assessment. Their prior work includes abundant literary translation projects, equipping them with the ability to handle typical cultural in-depth expressions of web novels. The third annotator is a student majoring in translation at a prestigious Chinese university, who participated both as an annotator and a quality supervisor. The student was responsible for performing annotation tasks while coordinating consistency checks and revising annotation guidelines based on feedback from the team.

The team followed a well-structured annotation workflow. Weekly annotation meetings were held to discuss challenging cases and refine annotation criteria. Feedback from translators was systematically integrated into guideline updates, enhancing both reliability and linguistic validity throughout dataset construction.

Together, the team's professional translation expertise, linguistic sensitivity, and collaborative workflow enabled the creation of a high-quality web novel translation benchmark. Their efforts ensured that the dataset is both linguistically faithful and contextually consistent, establishing a solid foundation for future research on evaluating LLM-based literary translation.

\subsection{Annotation Example}
We show some cases which can demonstrate our annotation guideline in Table \ref{tab:eval_examples}.

\subsection{Annotation Process}
\begin{figure*}[h] 
    \centering
    \includegraphics[width=0.9\linewidth]{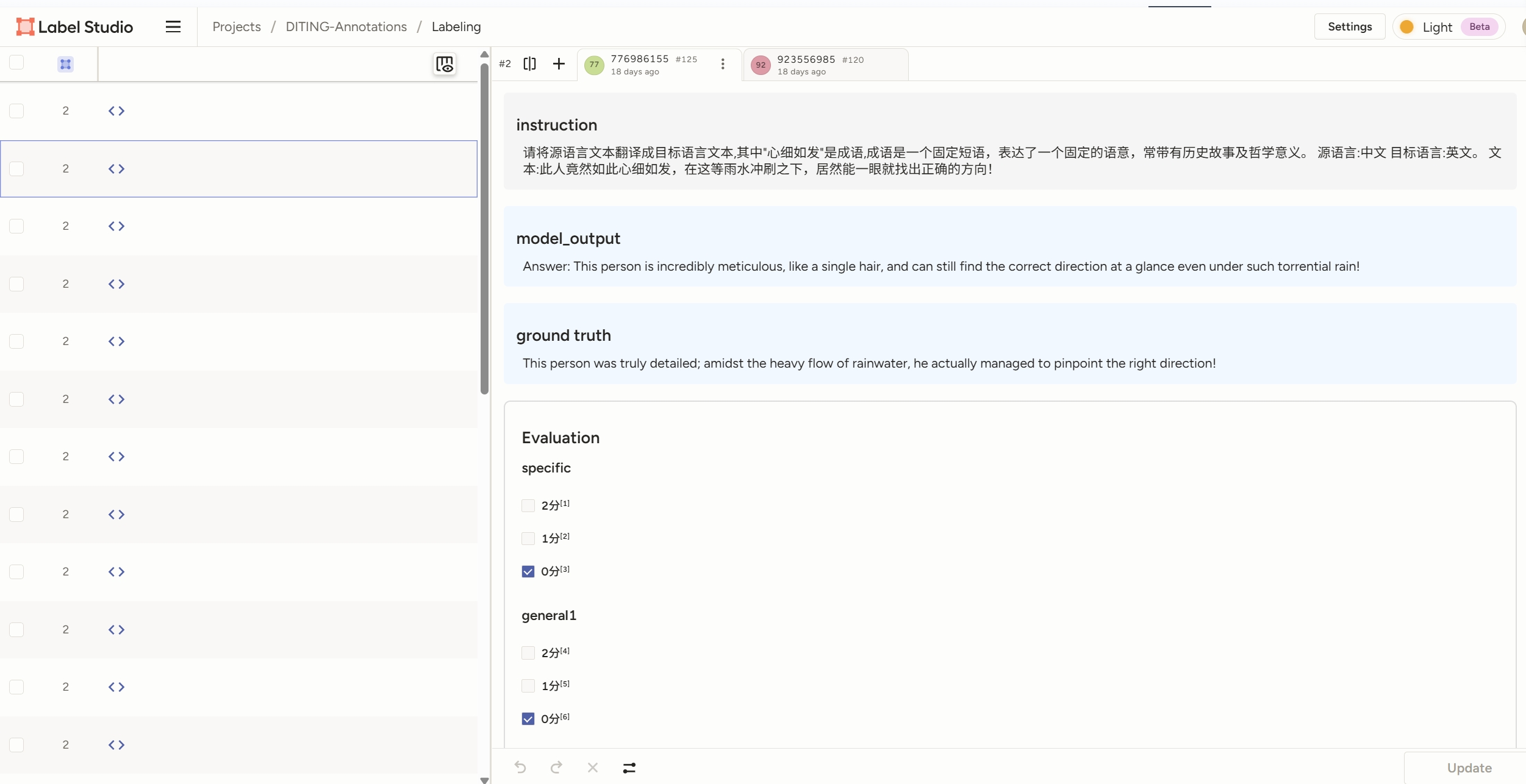}
    \caption{The Label Studio interface of the DITING annotation process.}
    \label{fig:annotation_process}
\end{figure*}

Our annotation process can be seen in Figure \ref{fig:annotation_process}.

\section{Experimental Details}
\label{Experiments}
\subsection{Evaluated Translation Models}
\label{app:benchmark_models}
Frontier LLMs include industry-leading APIs: 
\begin{itemize}
    \item GPT-4o, known for its strong general-purpose multilingual and reasoning abilities;
    \item DeepSeek-V3, which is optimized for Chinese and English tasks with enhancements in coding and mathematics;
\end{itemize}
For open-source models, we select large-scale foundation models:
 \begin{itemize}
     \item {DeepSeek-R1-70B}, which offers strong performance in complex reasoning;
     \item {LLaMA3-70B}, Meta's top-tier open-source model;
     \item {ChatGLM4-9B}, optimized for dialogue scenarios; 
     \item Qwen3 series, including Qwen3-8B, {Qwen3-14B}, and {Qwen3-32B};
     \item {LLaMA3-8B}, which balances efficiency and capability.
 \end{itemize}
Additionally, translation-specialized models are incorporated, including:
\begin{itemize}
    \item {Google Translate}, a widely-used commercial machine translation service based on large-scale neural methods;
    \item {IFLYTEK Translate}, a leading Chinese-oriented translation system;
    \item ByteDance's {Seed-X-PPO-7B}, fine-tuned with reinforcement learning;
    \item {Seed-X-Instruct-7B}; instruction-tuned for translation;
    \item  Xiaomi's {GemmaX2-28-9B}.
\end{itemize}
This selection ensures a diverse and representative evaluation across model types, scales, and specializations.

\subsection{Evaluation of Automatic Metrics}
\label{models}

We provide the detailed configuration of all the metrics used in our metrics Evaluation experiment in Table ~\ref{tab:metric_eval_settings}.

\subsection{Case Study}
We demonstrated how a case was evaluated within our framework in Table \ref{tab:good_case_study}.
\subsection{Prompts}
We show our prompts used in the evaluation framework which can be seen in Table \ref{tab:prompt}, Table \ref{tab:prompt2}.

\begin{table*}[ht]
\centering
\small
\renewcommand{\arraystretch}{1.3}
\caption{Representative Human Evaluation Examples Across Six Dimensions}
\label{tab:eval_examples}
\begin{tabular}{p{2cm} p{3.8cm} p{4.5cm} p{5cm}}
\toprule
\textbf{Dimension} & \textbf{Source Sentence (Chinese)} & \textbf{Example Translation (English)} & \textbf{Evaluation Comments} \\
\midrule

\textbf{Lexical Ambiguity} &
张骆宇感觉自己继续待在这里，也只不过是吃狗粮，还不如赶紧离开了。 &
\textit{“Zhang Luoyu felt that if he stayed here any longer, he would just be watching others show affection—it would be better for him to leave quickly.”} &
Specific Metric: \textbf{2} “watching others show affection” accurately conveys the idiomatic sense of “吃狗粮.” \newline
General 1: \textbf{2} Natural and idiomatic. \newline
General 2: \textbf{2} Fully preserves source meaning. \\
\addlinespace

\textbf{Terminology Localization} &
就是传说之中66年的拉菲、56年的飞天茅台这些一瓶价值数百万的珍贵名酒在这里也和其他普通酒一样，1生存币就能够购买一大杯。 &
\textit{“Even the legendary 1966 Lafite and the 1956 Feitian Moutai, which are precious wines worth millions of yuan per bottle, are sold here just like ordinary wines.”} &
Specific Metric: \textbf{1} “Feitian Moutai” should be rendered as “Flying Moutai”; numeric years may mislead. \newline
General 1: \textbf{1} Partial transliteration but inconsistent. \newline
General 2: \textbf{1} Slight redundancy. \\
\addlinespace

\textbf{Idiom Translation} &
此人竟然如此心细如发，在这等雨水冲刷之下，居然能一眼就找出正确的方向！ &
\textit{“This person is incredibly meticulous, like a single hair, and can still find the correct direction at a glance even under such torrential rain!”} &
Specific Metric: \textbf{0} Literal rendering of “如发” inappropriate; should use “meticulous.” \newline
General 1: \textbf{0} Idiomatic meaning lost. \newline
General 2: \textbf{0} Expression unnatural. \\
\addlinespace

\textbf{Zero Pronoun Translation} &
是的，自己被一个死神一般的男子救了下来，而且还答应了他必须接受他的要求！ &
\textit{“Yes, she was rescued by a man who seemed like a grim reaper, and agreed to his demands on the condition that he must accept her requirements.”} &
Specific Metric: \textbf{0} Clause logic incorrect; meaning deviates from source. \newline
General 1: \textbf{0} Pronoun reference wrong. \newline
General 2: \textbf{0} Illogical phrasing. \\
\addlinespace

\textbf{Tense Consistency} &
身高186cm的楚衍高出夏欣怡一头，他的声音在夏欣怡上方掠过。 &
\textit{“Chu Yanyan, who was 186 cm tall, stood a head taller than Xia Xinyi, his voice sweeping over her.”} &
Specific Metric: \textbf{2} Tense consistent and logical. \newline
General 1: \textbf{2} Fluent and aligns with English time conventions. \newline
General 2: \textbf{2} Slightly awkward unit usage (“six feet” preferred). \\
\addlinespace

\textbf{Cultural Safety} &
你TMD有完没完了，我两个月前和你上的床，你怀孕多久了。 &
\textit{“The hell are you still going on about this? I slept with you two months ago, how long have you been pregnant?”} &
Specific Metric: \textbf{1} Profanity softened appropriately. \newline
General 1: \textbf{0} Still carries offensive tone. \newline
General 2: \textbf{1} Sensitive content handled better than source. \\
\addlinespace

\bottomrule
\end{tabular}

\vspace{3pt}
\footnotesize{\textit{Note:} Each example shows a representative case of human evaluation across six task dimensions: lexical ambiguity, terminology localization, idiom translation, zero pronoun resolution, tense consistency, and safety. 
Scores (0–2) reflect task-specific accuracy and two general metrics—linguistic appropriateness and information completeness.}
\end{table*}

\begin{table*}[t]
   \caption{\label{tab:good_case_study} Case study. The table shows the relevant information about the evaluation of a case in our framework}
    \centering
    \small
    \renewcommand{\arraystretch}{1.3}
    \begin{tabular}{lp{13cm}}
    \toprule
    \multirow{1}{*}{Source} & {此 人 竟 然 如 此 心 细 如发，在这等雨水冲刷之
下，居然能一眼就找出
正确的方向！} \\
    \hline
    \multirow{2}{*}{Translation} & This person is as meticulous as a hair, and under such heavy rain, he can still find the correct direction at a glance!
  \\
    \hline
    \multirow{1}{*}{Gold Annotations} &Specific Metric: 0. Literal ren
dering of “如发” inappropriate,should use “meticulous.”

General Metric1: 0. Idiomatic meaning
lost.

General Metric2: 0. Expression unnatural.\\
\midrule
\multirow{1}{*}{Debate Round 1} &
\textbf{Agent1}:

Specific Metric: 0. General Metric1: 0. General Metric2: 0. 

The idiom “心细如发” is translated as "as meticulous as a hair", which is an inappropriate translation, resulting in a score of 0 for the specific indicator and thus a score of 0 for the general indicator.

\textbf{Agent2}:

Specific Metric: 0. General Metric1: 0. General Metric2: 0. 

The idiom "心细如发" (heart as meticulous as a hair) was translated as "as meticulous as a hair," an inappropriate translation.
This resulted in a score of 0 for this specific indicator, and consequently, a score of 0 for the overall indicator.

\textbf{Judge Agent}:

Judge:yes.

Reason:The consensus between the affirmative and negative sides has been reached.

\\
    \hline
     \multirow{1}{*}{Debate Progress} &
     END
        \\

     \hline
     \multirow{1}{*}{Model Evaluation} &Specific Metric: 0.
     
        General Metric1: 0. 
        
        General Metric2: 0.
        \\
    \hline

    \end{tabular}
 
\end{table*}

\begin{table*}[h]
\centering
\small
\renewcommand{\arraystretch}{1.3}
\caption{Translation Evaluation Task Instruction for Debater}
\label{tab:prompt}
\setlength{\fboxrule}{1pt} 
\setlength{\fboxsep}{10pt} 
\fbox{
\begin{tabular}{p{15cm}} 
You are a translation expert.\\

Translation task:\\
\textcolor{gray}{\textit{\{Task Details\}}}\\

Description of restrictive conditions:\\
\textcolor{gray}{\textit{\{Task Constraints，including response format and others\}}}\\

Here are the points you must pay attention to:\\
\textcolor{gray}{\textit{\{Some Notes about evaluation task\}}}\\

The following are the scoring rules. The scoring cases in them are for your reference:\\
\textcolor{gray}{\textit{\{The matching metrics\}}}\\

Scoring example:\\
\textcolor{gray}{\textit{\{Some gold annotations to refer\}}}\\

Davate progress:\\
\textcolor{gray}{\textit{\{The memory of debate progress\}}}

\end{tabular}
}
\end{table*}

\begin{table*}[h]
\centering
\small
\renewcommand{\arraystretch}{1.3}
\caption{Translation Evaluation Task Instruction for Judge}
\label{tab:prompt2}
\setlength{\fboxrule}{1pt} 
\setlength{\fboxsep}{10pt} 
\fbox{
\begin{tabular}{p{15cm}} 
You are a translation evaluation judge.\

Translation task:\\
\textcolor{gray}{\textit{\{Task Details\}}}\\

Description of restrictive conditions:\\
\textcolor{gray}{\textit{\{Task Constraints，including response format and others\}}}\\

Here are the points you must pay attention to:\\
\textcolor{gray}{\textit{\{Some Notes about evaluation task\}}}\\

The following are the scoring rules. The scoring cases in them are for your reference:\\
\textcolor{gray}{\textit{\{The matching metrics\}}}\\

Scoring example:\\
\textcolor{gray}{\textit{\{Some gold annotations to refer\}}}\\

Davate progress:\\
\textcolor{gray}{\textit{\{The memory of debate progress\}}}

Judge instructions:\

Determine whether the debating agents have reached a consensus.

If consensus is reached, record the agreed-upon evaluation.

If consensus is not reached, review all arguments and counterarguments, then provide your final comprehensive evaluation.

\end{tabular}
}
\end{table*}



\begin{table*}[htbp]
\centering
\caption{ Related configurations for experiments to verify the effectiveness of previous metrics.}  
\label{tab:metric_eval_settings}
\begin{tabular}{l l l}
\toprule
\textbf{Metric} & \textbf{Base / Tokenizer Model} & \textbf{Notes} \\
\midrule
BLEU & N/A & n-gram based metric, formula-based \\
ROUGE & N/A & recall-oriented n-gram metric, formula-based \\
ChrF & N/A & character n-gram based metric, formula-based \\
BLEURT & BleurtSPTokenizer & tokenizer class \\
COMET & xlm-roberta-large & base pretrain model \\
COMETkiwi-da & microsoft/infoxlm-large & base pretrain model \\
M-MAD & gpt-4o-mini (API) & base LLM \\
AgentEval$_{Ds-R1}$ & Deepseek-R1 (API) & base LLM \\
AgentEval$_{Debate-R1}$ & Deepseek-R1 (API) & base LLM \\
\bottomrule
\end{tabular}
\end{table*}

\end{CJK}

\end{document}